\documentclass[final,3p,times]{elsarticle}
\usepackage{bbding}
\usepackage{times}
\usepackage{enumitem}
\usepackage{epsfig}
\usepackage{graphicx}
\usepackage{amsmath}
\usepackage{algorithmic}
\usepackage{algorithm}
\usepackage{amssymb}
\usepackage{epstopdf}
\usepackage{subfigure}
\usepackage{url}
\usepackage{multirow}
\graphicspath{{fig/}}

\usepackage{adjustbox}
\usepackage{amssymb}
\begin{document}

\begin{frontmatter}



\title{Collaborative Teacher-Student Learning via Multiple Knowledge Transfer}

\author[1]{Liyuan Sun}
\author[1]{Jianping Gou}
\ead{goujianping@ujs.edu.cn}
\address[1]{School of Computer Science and Telecommunication Engineering, Jiangsu University, Zhenjiang,
212013, China}

\author[3]{Baosheng Yu}

\author[2]{Lan Du}

\author[3]{Dacheng Tao}

\address[2]{Faculty of information technology, Monash University, Australia}
\address[3]{UBTECH Sydney AI Centre, School of Computer Science, Faculty of Engineering, The University of Sydney, Darlington, NSW 2008, Australia.}

\begin{abstract}
Knowledge distillation (KD), as an efficient and effective model compression technique, has been receiving considerable attention in deep learning.~The key to its success is to transfer knowledge from a large teacher network to a small student one.~However, most of the existing knowledge distillation methods consider only one type of knowledge
learned from either instance features or instance relations via a specific distillation
strategy in teacher-student learning.~There are few works that explore the idea of transferring different types of knowledge with different distillation strategies in a unified framework.~Moreover, the frequently used offline distillation suffers from a limited learning capacity due to the fixed teacher-student architecture.
In this paper we propose a collaborative teacher-student learning via multiple knowledge transfer (CTSL-MKT)
that prompts both self-learning and collaborative learning.~It allows multiple students learn knowledge from both individual instances and instance relations in a collaborative way.~While learning from themselves with self-distillation, they can also guide each other via online distillation.~The experiments and ablation studies on four image datasets
demonstrate that the proposed CTSL-MKT significantly outperforms the state-of-the-art KD methods.
\end{abstract}


\end{frontmatter}

% \linenumbers

\section{Introduction}

Deep neural networks have achieved state-of-the-art performance on many applications such as computer vision, natural language processing, and speech recognition in recent years.~The remarkable performance of deep learning relies on designing deeper or wider network architectures with many layers and millions of parameters to enhance the learning capacity.~However, it is almost impossible to deploy the large-scale networks on platforms with limited computation and storage resources,  {\it e.g.}, mobile devices and embedded systems.~Thus, the model compression and acceleration techniques mainly including network pruning \cite{li2017,han2015}, model quantization \cite{rastegari2016,hubara2016} and knowledge distillation \cite{hinton2015,romero2015,xu2019,cho2019,chen2020} are proposed for training lightweight deep models.~Among compressing methods, knowledge distillation, which carries out knowledge transfer from a high-capacity teacher network to a low-capacity student one, has received increasing interest recently since it was first introduced in  \cite{hinton2015}.

In knowledge distillation, the type of knowledge, the distillation strategy and the teacher-student architecture are three crucial factors that determine the KD performance \cite{Gou2020}.~As pointed out in \cite{Gou2020}, there are three kinds of knowledge, {\it i.e.}, the response-based, the feature-based and the relation-based knowledge.~Generally, most of KD methods distill the response-based knowledge ({\it e.g.}, soft logits of the output layer) from a large teacher network and transfer it to a small student \cite{hinton2015,kim2019, mirzadeh2019}.~To overcome the limitation of knowledge from the output layer of teacher, the feature-based knowledge from the middle layers of teacher is
also used to train the student \cite{romero2015,XuK2020,Guan2020}. Unlike both the response-based and the feature-based knowledge from individual instances, the relation-based knowledge from instance relations is modelled for improving student learning \cite{you2017,wu2020embedding,Yim2017,Passalis2020a,YuYazici2019}.
Each kind of knowledge can provide student training with an informative teacher guidance, and they can also compensate each other to enrich learning.
However, most existing KD methods only consider either knowledge from individual instance features to maintain instance consistency between teacher and student or knowledge from instance relations to preserve the instance correlation consistency.
There are a few works that consider more than one kind of knowledge in knowledge distillation \cite{you2017,ShenCX2019} at the same time and explore the efficacy of each kind of knowledge.

Transferring different types of knowledge can be implemented with different distillation methods, {\it e.g.}, offline distillation, online distillation and self-distillation \cite{Gou2020}.~Most of the KD methods employ offline distillation, which is one-way knowledge transfer from a pre-trained large teacher to a small student \cite{hinton2015,mirzadeh2019,LiLi2018}.~In offline distillation, the capacity gap caused by a fixed teacher-student architecture and the requirement of a large dataset for pre-training the teacher often
result in a degraded performance~\cite{mirzadeh2019}.~Thus, finding a proper teacher-student architecture in offline distillation
is challenging.~In contrast, online distillation provides a one-phase end-to-end training scheme via teacher-student collaborative learning on a peer-network architecture instead of a fixed one \cite{zhang2018,hou2017,chen2020,kim2019,lan2018,Walawalkar2020}. Self-distillation performs online distillation within the same network to reduce model over-fitting \cite{yuan2020,zhang2019}.~Online distillation and self-distillation are promising methods for knowledge distillation as they bridge
the capacity gap via avoiding the need of a large teacher network, leading to an improved performance.
However, both KD methods used individually are limited to knowledge distillation from a single source, {\it i.e.}, individual instances,
online distillation could further suffer from the poor instance consistency between peer networks caused by the discrepancy in their network outputs.

\begin{figure}[!t]
\centering
\includegraphics[scale=0.8]{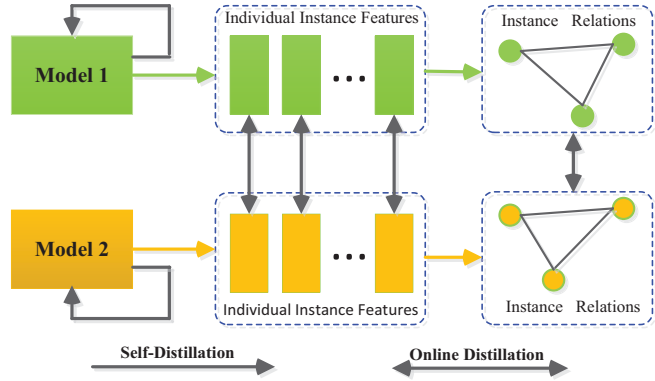}
\caption{The overview diagram of CTSL-MKT.}
\label{fig1}
\end{figure}

Consequently, it is desirable to have a unified framework that can integrate the advantages of different KD methods and make
efficient use of different types of knowledge.
Inspired by the idea of knowledge distillation via multiple distillation strategies to transfer more than one types of knowledge,
we propose a collaborative teacher-student learning via multiple knowledge transfer (CTSL-MKT), which fuses self-distillation and online distillation in such a way that the former transfers the response-based knowledge within each peer network and
the latter bidirectionally transfers both the response-based knowledge and the relation-based knowledge between peer networks.~CTSL-MKT can overcome the aforementioned issued faced by existing KD methods that often use only one distillation strategy to transfer a single type of knowledge.
The overview framework of CTSL-MKT is illustrated in Figure~\ref{fig1}.
To our knowledge, this is the first framework that integrates different distillation strategies together
to transfer more than one type of knowledge simultaneously.

In CTSL-MKT, each pear network conducts self-learning via self-distillation.~Meanwhile, they carry out teacher-student collaborative learning to mutually teach each other.~CTSL-MKT can also adopt a variety of peer network architectures, where the two peer
networks can either share the same network architecture or have different ones.~We believe that multiple knowledge transfer can provide much more informative knowledge to guide each peer
network so that they can obtain better performance with a better generalization ability.
We conduct a set of image classification experiments on four commonly-used datasets {\it i.e.},~CIFAR-10, CIFAR-100, Tiny-ImageNet, and Market-1501.~Experimental results demonstrate the superior performance of the proposed CTSL-MKT over the state-of-the-art KD methods.
The main contributions in our works can be summarized as follows:
\begin{itemize}[itemsep=0pt,parsep=0pt]
\item A new teacher-student mutual learning framework effectively fuses the knowledge from individual instances and the knowledge from instance relationships.
\item A self-learning enhanced collaborative learning integrates the advantages of both self-learning and online learning.
\item The extensive experiments on a variety of the peer teacher-student networks that compare CTSL-MKT with the state-of-the-art methods to validate its effectiveness in image classification tasks.
\item A set of ablation studies of different combinations of knowledge and distillation methods provides insights into how multiple knowledge transfer contribute to knowledge distillation.
\end{itemize}

\section{Related Work}

\subsection{Self-Distillation}

Self-distillation is a novel training scheme for knowledge transfer \cite{yuan2020,zhang2019,yun2020,Phuong2019,YangCXie2019}.~In self-distillation, the teacher and student networks are identical and knowledge transfer is carried out within the same network.~Yuan {\it et al.}~empirically analyzed the performance of normal, reversed and defective KD methods, and showed that a weak teacher can strengthen the student and vice-versa \cite{yuan2020}.~A teacher-free knowledge distillation method (Tf-KD) instead makes student model conduct self-learning.~To enhance the generalization and overcome over-fitting, class-wise self-knowledge distillation makes use of soft logits of different intra-class samples within a model~\cite{yun2020}.~Phuong and Lampert \cite{Phuong2019} proposed a distillation-based training method to reduce time complexity,  where the output of later exit layer supervises the early exit layer via knowledge transfer.
Rather than at the layer level, snapshot distillation \cite{YangCXie2019}
transfers knowledge from earlier to later epochs while training a deep model.

Overall, self-distillation can overcome the issue of over-fitting and the capacity gap on the teacher-student architectures, improve the generalization ability and reduce the inference time of a deep model.~However, the self-distillation performance could be limited by the one-sided response-based knowledge from model itself.~To further improve knowledge distillation, we integrate both online and self-distillation into CTSL-MKT with more informative relation-based knowledge.

\subsection{Collaborative Learning}

Recently, there are many new online distillation methods that train a teacher and a student simultaneously during knowledge transfer.
Collaborative learning is the one used most often  \cite{zhang2018,wu2020,kim2019,YaoA2020,Guo2020,LiK2020},
where the teacher and the student as peer networks collaboratively teach and learn from each other, and the peer network architectures can be different.~In particular, Zhang {\it et al.} \cite{zhang2018} proposed a deep mutual learning method (DML) for online distillation using the response-based knowledge.~DML uses an ensemble of soft logits as knowledge and transfers it among arbitrary peer networks via collaborative learning~\cite{Guo2020}.~Yao and Sun \cite{YaoA2020} further extended DML with dense cross-layer mutual-distillation, which learns both the teacher and the student collaboratively from scratch.

Unlike the ensemble of peer networks, the advantage of a mutual knowledge distillation method is that it can fuse features of peer networks to collaboratively learn a powerful classifier~\cite{kim2019}.~However, the knowledge distilled by those online mutual distillation methods is limited to the response-based knowledge from individual instance features.
In contrast, our work can further make use of the relation-based knowledge from instance relationships to further
enrich the transferred knowledge.

% These methods argued above can be summarily called online mutual distillation, which can strengthen generalization ability of deep models, save computational time and achieve the satisfactory performance. However, online mutual distillation almost uses the response-based knowledge from individual instance features, so as to cause the instance inconsistency for the same individual samples and ignore instance correlation consistency. In contrast to them, our work introduces the relation-based knowledge from instance relationships and unifies self-distillation into online distillation.

\subsection{Structural Knowledge}

Most of knowledge distillation approaches adopt the output logits of a deep model from individual samples as knowledge and make the logits of the teacher and student match each other.~However, such response-based knowledge ignores the structural knowledge from the mutual relations of data examples,  known as relation-based knowledge.~In recent years, there are some newly proposed knowledge distillation methods based on structural relations of data samples \cite{park2019,liu2019,peng2019,LiuY2020,Tung2019,ChenH2018}. Park {\it et al.}~\cite{park2019} proposed a relational knowledge distillation method (RKD), which transfers the instance relation knowledge from a teacher to a student.~Chen {\it et al.}~\cite{ChenH2018} borrowed the idea of manifold learning to design a novel knowledge distillation method, in which the student preserves the feature embedding similarities of samples from the teacher.~Peng {\it et al.}~\cite{peng2019} designed a knowledge transfer method that makes sure the student matches the instance correlation consistently with the teacher.

However, those structural knowledge distillation methods often ignore the knowledge directly from individual samples.
Our proposed CTSL-MKT instead considers the knowledge from both individual instances and instance relationships,
and the bidirectional knowledge transfer is carried out between peer networks via collaborative learning.

% Most of knowledge distillation approaches adopt the output logits of deep model from individual samples as knowledge and make the logits of the teacher and student match each other. However, such response-based knowledge ignores the structural knowledge from the mutual relations of data examples, which is a form of relation-based knowledge.~In recent years, there are some newly proposed knowledge distillation methods based on structural relations of data samples \cite{park2019,liu2019,peng2019,LiuY2020,Tung2019,ChenH2018}. To be specific, Park {\it et al.} propose a relational knowledge distillation method (RKD), which transfers the instance relation knowledge from teacher to student \cite{park2019}. Chen {\it et al.} borrow the idea of manifold learning to design a novel knowledge distillation method, in which the student preserves the feature embedding similarities of samples from teacher~\cite{ChenH2018}.  Peng {\it et al.} design knowledge transfer to make student match the instance correlation congruence from teacher~\cite{peng2019}.

% However, those structural knowledge distillation methods almost transfer knowledge via offline distillation and often ignore the knowledge directly from individual samples. To address the issues, our proposed CTSL-MKT simultaneously considers the knowledge from both individual instances and instance relationships, and the bidirectional knowledge transfer is carried out between peer networks via collaborative learning.

\section{The Proposed CTSL-MKT Method}
\label{3}

\begin{figure*}[!t]
\centering
\includegraphics[scale=0.95]{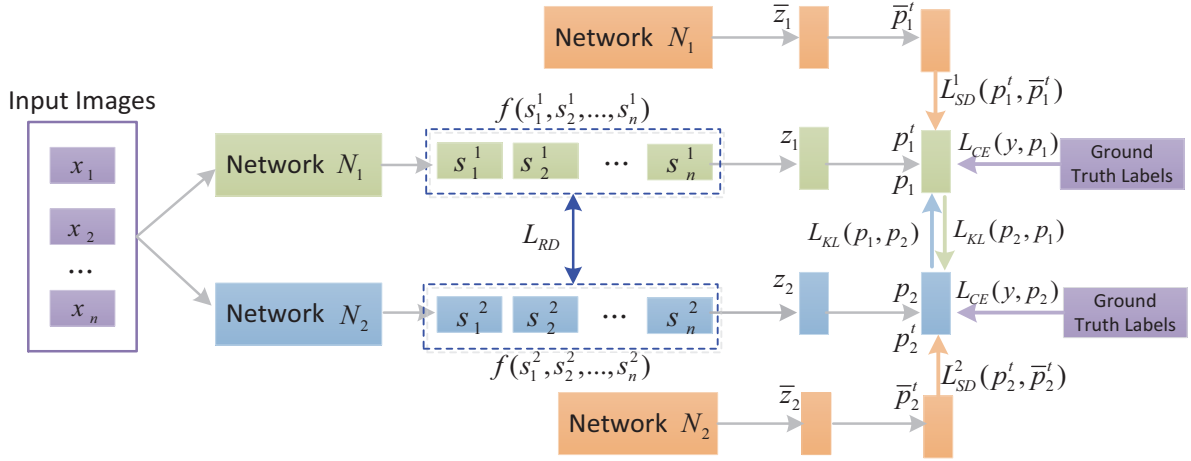}
\caption{The framework of CTSL-MKT with two peer networks. Note that the losses $L_{KL}(\emph{\textbf{p}}_1,\emph{\textbf{p}}_2)$ and $L_{KL}(\emph{\textbf{p}}_2,\emph{\textbf{p}}_1)$ are for mutual learning via the response-based knowledge transfer, $L_{RD}$ for mutual learning via the relation-based knowledge transfer, $L_{SD}^k(\emph{\textbf{p}}_k^{t},\bar{\emph{\textbf{p}}}_k^{t})$ for self-learning via the response-based knowledge transfer.}
\label{fig2}
\end{figure*}

CTLS-MKT unifies student self-learning and teacher-student mutual learning under one framework in such a way that it can utilise multiple types of knowledge and more than one distillation methods during the teacher-student learning.~The teacher-student architectures used in CTSL-MKT are peer networks, such as ResNet \cite{he2016deep} and MobilenetV2 \cite{sandler2018mobilenetv2}.~Different from previous works, CTSL-MKT distills both the response-based knowledge from individual instance features and the relation-based knowledge from instance relationships.~During teacher-student mutual learning, peer networks trained collaboratively can teach each other via online distillation with the two kinds of knowledge.~Meanwhile, each peer network can also self-learn via self-distillation with the response-based knowledge.~The two leaning processes working together can complement each other to explore different knowledge space and enhance the learning.~Our CTSL-MKT can be seen as a new model compression technique that generalises the existing self-distillation and online distillation methods and
enables fast computations and improves the generalization ability.
The overall framework of CTSL-MKT with two peer networks is shown in Figure~\ref{fig2}
as an example, and notations are summarized in Table~\ref{tb0}.

\begin{table}[!t]%\footnotesize
\centering
\begin{adjustbox}{max width=\linewidth}
\begin{tabular}{lp{0.7\linewidth}}
\hline
    Notations & Descriptions\\ \hline
$X=\{x_1, x_2,\cdots, x_n\}$  & $n$ input samples from $m$ classes \\
$\emph{\textbf{y}}=\{y^{1}, y^{2},\cdots, y^{m}\}$  & the one-hot label vector for $x\in X$\\
%$N_{k}$  & deep network $k$\\
$\emph{\textbf{z}}_k(x)=\{z_k^1,z_k^2,\ldots,z_k^m\}$ & logits of a network $N_{k}$ for $x\in X$ where $z_k^{i}$ is the logit for class $i$\\
%$t$  & temperature parameter\\
$\sigma_i(\emph{\textbf{z}}_k(x),t)$  & softmax function with temperature $t$\\
%$\sigma_i(\emph{\textbf{z}}_k(x),1)$  & softmax function when $t=1$\\
$p_k^{it}=\sigma_i(\emph{\textbf{z}}_k(x),t)$  & output of softmax for $z_k^{i}$ \\
%$p_k^{i}=\sigma_i(\emph{\textbf{z}}_k(x),1)$  & output of softmax for $z_k^{i}$ \\
$\emph{\textbf{p}}_k^{t}=\{p_k^{1t},p_k^{2t},\ldots,p_k^{mt}\}$  & predictions of $N_{k}$ with $t$ \\
$\emph{\textbf{p}}_k=\{p_k^{1},p_k^{2},\ldots,p_k^{m}\}$  & predictions of $N_{k}$ when $t=1$ \\
%$\phi(.)$  & feature mapping function of $N_{k}$ \\
%$L_{KL}(.)$    &  Kullback-Leibler divergence \\
%$s_i^k=\phi(x_i)$  & output of any layer of $N_{k}$ for $x_i$ \\
$f(s_1^k, s_2^k,\cdots, s_n^k)$  & similarity loss of $n$ samples in $N_{k}$\\
%$R(.)$  & instance relation loss\\
\hline
\end{tabular}
\end{adjustbox}
\caption{Notations used in CTSL-MKT.}
\label{tb0}
\end{table}

% \subsection{The CTSL-MKT Overview}
% \label{3.1}

% To explore the efficacy of knowledge distillation with multiple types of knowledge and more than one distillation strategies during the teacher-student learning, the proposed CTSL-MKT includes student self-learning and teacher-student mutual learning with two kinds of knowledge and two distillation schemes. The teacher-student architectures in CTSL-MKT are peer networks. The knowledge contains response-based knowledge from individual instances and relation-based knowledge from instance relationships. To be specific, during teacher-student mutual learning peer networks are collaboratively trained and teach each other by online distillation with two kinds of knowledge, and simultaneously each peer network is trained by self-distillation with response-based knowledge during student self-learning. The CTSL-MKT framework with its model formulations is shown in Fig.~\ref{fig2}.

% Through collaborative learning via multiple knowledge transfer in CTSL-MKT, multiple kinds of knowledge can provide the informative knowledge, peer networks easily form a variety of teacher-student architectures; multiple distillation strategies can transfer knowledge as far as possible. Thus,

\subsection{Teacher-Student Mutual Learning}
\label{3.2}
Teacher-student mutual learning contains the response-based knowledge transfer and the relation-based knowledge transfer among peer network architectures.
%For easy understanding, we demonstrate the idea of mutual learning with two peer networks $N_{1}$ and $N_{2}$ below.

\textbf{Response-Based Knowledge Transfer}:
% \label{3.2.1}
The response-based knowledge ({\it i.e.}, the output of a peer network) is learned from individual instances.~Given a peer network $N_k$ and its output $\emph{\textbf{p}}_k$ with temperature parameter $t=1$,
the collaborative response-based knowledge transfer makes the student network $N_{k}$ imitate the teacher network $N_{k'}$ ($k \neq k'$)
with the following Kullback-Leibler (KL) divergence loss,
\begin{equation}\label{eq3}\small
    L_{KL}(\emph{\textbf{p}}_k,\emph{\textbf{p}}_{k'})=\sum_{x\in X}\sum_{i=1}^m\sigma_i(\emph{\textbf{z}}_{k'}(x),1)log\frac{\sigma_i(\emph{\textbf{z}}_{k'}(x),1)}{\sigma_i(\emph{\textbf{z}}_{k}(x),1)}.
\end{equation}
Similarly, the loss that the student network $N_{k'}$ uses to learn from the teacher network $N_{k}$ is $L_{KL}(\emph{\textbf{p}}_{k'},\emph{\textbf{p}}_k)$.

During the collaborative learning for a classification task, each peer network $N_{k}$ will then be trained
with both the KL divergence loss (Eq~(\ref{eq3})) and the cross-entropy (CE) loss (Eq~(\ref{eq2})).
\begin{equation}\label{eq2}
    L_{CE}(\emph{\textbf{y}},\emph{\textbf{p}}_k)=-\sum_{x\in X}\sum_{i=1}^m y^{i}log(\sigma_i (\emph{\textbf{z}}_k(x),1))~.
\end{equation}
Take two peer networks in Figure~\ref{fig2} as example,
the losses used to train $N_1$ and $N_2$ will be
$L_{CE}(\emph{\textbf{y}},\emph{\textbf{p}}_1)+L_{KL}(\emph{\textbf{p}}_1,\emph{\textbf{p}}_2)$
and $L_{CE}(\emph{\textbf{y}},\emph{\textbf{p}}_2)+L_{KL}(\emph{\textbf{p}}_2,\emph{\textbf{p}}_1)$, respectively.

% Of course, the cross-entropy loss for training network $N_{2}$ is $L_{CE}(\emph{\textbf{y}},\emph{\textbf{p}}_2)$.
% Thus, during collaborative learning, the peer networks $N_{1}$ and $N_{2}$ are learned with the following two losses, $L_{CE}(\emph{\textbf{y}},\emph{\textbf{p}}_1)+L_{KL}(\emph{\textbf{p}}_1,\emph{\textbf{p}}_2)$ and $L_{CE}(\emph{\textbf{y}},\emph{\textbf{p}}_2)+L_{KL}(\emph{\textbf{p}}_2,\emph{\textbf{p}}_1)$, respectively.

\textbf{Relation-Based Knowledge Transfer}:~CTSL-MKT further integrates the relation-based knowledge learned from the instance relationships via the teacher-student mutual leaning
in order to enrich the transferred knowledge and enhance the teacher guidance.~Let $s_j^k=\phi_k(x_j)$ (where $\phi_k(.)$ is a feature mapping function of $N_{k}$)
be the output of any layer of the network $N_{k}$ for $x_j$,
and
$\chi^\tau$ denote a set of $\tau$-tuples of different samples.~A set of $2$-tuples and a set of $3$-tuples thus correspond to $\chi^2=\left\{(x_u,x_v)|u\ne v\right\}$
and $\chi^3=\left\{x_u,x_v,x_w)|u\ne v\ne w\right\}$, respectively. As in \cite{park2019}, the relation-based knowledge learned by the network $N_{k}$ can be modelled jointly by a distance-wise function and an angle-wise function.

Given $N_k$, the distance-wise function captures the similarities between two samples in a $2$-tuple, which is defined as
\begin{equation}\label{eq4}
   f(s_u^k, s_v^k )=\frac{1}{\pi}||s_u^k-s_v^k ||_2~,
\end{equation}
where $\pi=\frac{1}{|\chi^2|}\sum_{(x_u,x_v)\in \chi^2}||s_u^k-s_v^k||_2$ is a normalization constant.
Accordingly, the instance relationships between any two peer networks $N_{k}$ and $N_{k'}$ are transferred
by the following distance-wise distillation loss
\begin{equation}\label{eq6}
  L_{DD}(x_u,x_v)=\sum_{(x_u,x_v)\in \chi^2} R\big(f(s_u^{k}, s_v^{k}),f(s_u^{k'}, s_v^{k'})\big)~,
\end{equation}
where $R(.)$ is Huber loss that reflects instance relationships and is defined as
\begin{equation}\label{eq7}
R(a,b)=
\left\{
             \begin{array}{lr}
            \frac{1}{2}(a-b)^2,\quad   if~|a-b|\leq1&  \\
             |a-b|-\frac{1}{2},\quad otherwise &
             \end{array}
\right.~.
\end{equation}
Furthermore, the similarities between samples in a $3$-tuple are measured by an angle-wise function
\begin{equation}\label{eq8}
   f(s_u^k, s_v^k, s_w^k )=\cos\angle s_u^k s_v^k s_w^k=<e^{uv},e^{wv}>~,
\end{equation}
where $e^{uv}=\frac{s_u^k-s_v^k}{||s_u^k-s_v^k||_2}$ and $e^{wv}=\frac{s_w^k-s_v^k}{||s_w^k-s_v^k||_2}$.
The instance relationships are transferred between any two peer networks $N_{k}$ and $N_{k'}$ with the angle-wise distillation loss, defined as
\begin{equation}\label{eq9}
\begin{aligned}
  &L_{AD}(x_u, x_v, x_w)\\
  =&\sum_{(x_u, x_v, x_w)\in \chi^3}R\big(f(s_u^{k}, s_v^{k},s_w^{k}),f(s_u^{k'}, s_v^{k'}, s_w^{k'})\big)~.
\end{aligned}
\end{equation}

It has been shown that the relation-based knowledge transfer can be more effective
if the distance-wise function is used jointly with the angle-wise function \cite{park2019}, as they capture different degrees of similarities between samples.
We formulate the instance relation distillation loss used in the collaborative learning between peer networks as
\begin{equation}\label{eq10}
  L_{RD}=L_{DD}(x_u,x_v)+\beta_1 L_{AD}(x_u, x_v, x_w)~,
\end{equation}
where $\beta_1$ is a tuning parameter that controls the balance between loss terms.

Consequently,
the mutual distillation loss with both the response-based and the relation-based knowledge
between two peer networks ($N_k$ and $N_{k'}$) is defined as: for network $N_k$, we have
\begin{equation}\label{eq11}
  L_{MD}^k=L_{RD}+\beta_2 L_{KL}(\emph{\textbf{p}}_k,\emph{\textbf{p}}_{k'})~,
\end{equation}
where $\beta_2$ is a tuning parameter; for network $N_{k'}$, we have
\begin{equation}\label{eq112}
  L_{MD}^{k'}=L_{RD}+\beta_2 L_{KL}(\emph{\textbf{p}}_{k'},\emph{\textbf{p}}_k)~.
\end{equation}

\renewcommand{\algorithmicensure}{\textbf{Output:}}
\begin{algorithm}[!t] \footnotesize
\caption{The proposed CSL-MKT}\label{alg}
\begin{algorithmic}[1]
\REQUIRE
Input samples $X$ with labels, learning rate $\eta$, hyperparameters $\alpha$, $\beta$, $\gamma$, $\beta_1$ and $\beta_2$.
\STATE \textbf{Initialize:} Initialize peer networks $N_1$ and $N_2$ to different conditions.
\STATE \textbf{Stage 1:}  Pre-train $N_1$ and $N_2$ for use of the process of self-learning.
\FOR{k=1 to 2}
\STATE \textbf{Repeat:}
\STATE  Compute stochastic gradient of $L_{CE}$ in Eq.~\eqref{eq2} and update $N_k$:
\STATE \ \ \ \ \ \ \ \ \ \ \ \ \ \ \ \ \ \ \ \ \ \ \ \ $N_k\gets \text{$N_k$}$+$\eta$$\frac{\partial L_{CE}}{\partial N_k}$.
\STATE \textbf{Until:} $L_{CE}$  converges.
\ENDFOR
%\STATE \textbf{end for}
\STATE \textbf{Stage 2:}  Train $N_1$ and $N_2$ collaboratively.
\STATE \textbf{Repeat:}
\FOR{k=1 to 2}
%\STATE  Compute $L_{ce}$ by ~\eqref{eq2} to calculate the cross entropy between the network output and the ground truth labels.
%\STATE  Compute $L_{mutual}$ by ~\eqref{eq11} to calculate the mutual learning loss function between the networks.
%\STATE  Compute $L_{cl}$ by ~\eqref{eq12} to regularize the network.
\STATE  Compute stochastic gradient of $L_{KD}^k$ in Eq.~\eqref{eq13} and update $N_k$:
%\STATE Compute total loss by ~\eqref{eq13} of the network.
\STATE \ \ \ \ \ \ \ \ \ \ \ \ \ \ \ \ \ \ \ \ \ \ \ \ $N_k \gets \text{$N_k$}$+$\eta$$\frac{\partial L_{KD}^k}{\partial N_k}$.
\ENDFOR
\STATE \textbf{Until:} $L_{KD}^k$ converges.
\STATE \textbf{return} \emph{$N_k$}.
\end{algorithmic}
\end{algorithm}

\subsection{Student Self-learning}
\label{3.3}

During the collaborative learning between peer networks, if the outputs of peer networks are very diverse, the mutual knowledge transfer could become poor.~Since the self-learning  via self-distillation can improve the power of knowledge transfer~\cite{yuan2020}, CTSL-MKT further introduces the self-learning of each peer network into collaborative learning via the response-based knowledge self-distillation. To conduct self-learning for each peer network $N_{k}$,
we use the outputs $\bar{\emph{\textbf{p}}}_k^{t}$ of the pre-trained network $N_{k}$ to supervise itself
with the following
self-distillation loss:
\begin{equation}\label{eq12}\small
L_{SD}^k(\emph{\textbf{p}}_k^{t},\bar{\emph{\textbf{p}}}_k^{t})=\sum_{x\in X}\sum_{i=1}^m\sigma_i^t(\bar{\emph{\textbf{z}}}_k(x),t)log\frac{\sigma_i^t(\bar{\emph{\textbf{z}}}_k(x),t)}{\sigma_i^{t}(\emph{\textbf{z}}_k(x),t)}~.
\end{equation}

\subsection{The CSL-MKT Algorithm}
\label{3.4}

Finally, CSL-MKT conducts mutual learning and self-learning simultaneously in a unified framework,
as shown in Algorithm~\ref{alg}.
Its objective function for each pear network is defined as
\begin{equation}\label{eq13}
  L_{KD}^k=\alpha L_{CE}^k+\beta L_{MD}^k+\gamma L_{SD}^k~,
\end{equation}
where $\alpha$, $\beta$ and $\gamma$ are the tuning parameters, which
balance the contribution of each loss in the collaborative learning,
$L_{CE}^k$ is defined in Eq.~\eqref{eq2}, $L_{MD}^k$ for two peer networks in Eqs.~\eqref{eq11} or \eqref{eq112}, and $L_{SD}^k$ in Eq.~\eqref{eq12}.

\section{Experiments}
\label{4}

\begin{table}[!t]\footnotesize
\centering
%\resizebox{.95\columnwidth}{!}{
\begin{tabular}{c|c|ccc}
\hline
    \multirow{2}{*}{Network} & \multirow{1}{*}{Parameter size} &\multicolumn{3}{c}{Baseline} \\
    \cline{3-5}
     &(CIFAR-100) &B\_10 &B\_100 &B\_Tiny\\
\hline
    ResNet14 &6.50M &94.94 &76.12&- \\
    ResNet18  &11.22M & 95.13 &75.77&62.90\\
    ResNet34   & 21.33M &95.39 & 77.66&- \\
    VGG19   & 139.99M &92.83 & 69.42&- \\
    MobileNetV2 & 2.37M &90.97  &68.23&- \\
    ShuffleNetV2 & 1.36M & 91.03 &70.10&- \\
    AlexNet &57.41M  & - &- &50.25\\
    SqueezeNet &0.77M  & - &-&43.68\\
\hline
\end{tabular}
\caption{The parameter size of each peer network on CIFAR-100 and its classification performance on three datasets.~Note that B\_10, B\_100, and B\_Tiny denote Top-1 accuracy (\%) achieved by each peer network on CIFAR-10, CIFAR-100 and Tiny-ImageNet, respectively.}
\label{tbbase}
\end{table}

We conducted extensive experiments to verify the effectiveness of CTSL-MKT on image classification tasks using datasets including CIFAR-10~\cite{krizhevsky2009learning}, CIFAR-100 \cite{krizhevsky2009learning}, Tiny-ImageNet \cite{deng2009imagenet} and Market-1501 \cite{zheng2015scalable}.~The peer network architectures were chosen from ResNet \cite{he2016deep}, MobileNet \cite{sandler2018mobilenetv2}, ShuffleNet \cite{ma2018shufflenet}, VGG \cite{simonyan2015very}, AlexNet \cite{krizhevsky2012imagenet} and SqueezeNet \cite{iandola2017squeezenet}.~CTSL-MKT was compared to the state-of-the-art KD methods, which are DML \cite{zhang2018}, Tf-KD \cite{yuan2020} and RKD \cite{park2019}.~For a fair comparison, RKD uses online distillation with the peer networks.~In all the experiments, the relation-based knowledge was modelled by the final feature embedding outputs by the peer networks.

\subsection{Datasets and Settings}
\label{4.0}

\noindent\textbf{CIFAR-10 and CIFAR-100.}
Both datasets have 60,000 $32\times 32$ images,
where 50,000 images are for training and the other 10,000 images are for testing.~The number of classes is 10 for CIFAR-10 and
100 for CIFAR-100,
and each class has the same numbers of samples in both the training and the testing sets.~On each dataset, data augmentation with random crops and horizontal flips was used to change the zero-padded $40\times 40$ images to $32\times 32$ ones.
The peer networks were trained for 200 epochs with batch size 128 and initial learning rate 0.1 which
is then multiplied by 0.2 at 60, 120, and 160 epochs.~Temperature parameter was set to 10 for CIFAR-10 and 3 for  CIFAR-100.

\noindent\textbf{Tiny-ImageNet.}~Tiny-ImageNet contains 100,000 training and 10,000 testing $64\times 64$ images from 200 classes, each of
which has the same number of samples.
Each image was randomly resized to $224\times 224$.~The peer networks were trained for 90 epochs with batch size 64 and initial learning rate 0.1 which is then multiplied by 0.1 at 30, 60, 80 epochs.~Temperature parameter was set to 2.

\noindent\textbf{Market-1501.} Market-1501 includes 32,688 images taken from 1,501 identities under condition of six camera views. It has 751 identities for training and 750 ones for testing. Each image was zero-padded by 10 pixels on each side, and data augmentation with random crops and horizontal flips was used to change the zero-padded images to $256\times 128$ ones.~The peer networks were trained for 60 epochs with batch size 32 and initial learning rate 0.05 which is then multiplied by 0.2 at 40 epochs. Temperature parameter was set to 6.

We used the SGD optimizer for training the peer networks with momentum 0.9 and weight decay 5e-4, and all input images were normalized by Mean-Std normalization.~All the hyper-parameters were greedily searched and set as follows.~The hyper-parameters used on CIFAR and Tiny-ImageNet were set as $\alpha=0.1$, $\beta=0.05$ and $\gamma=0.9$ for MobileNet and ShuffleNet, and $\alpha=0.4$, $\beta=0.4$ and $\gamma=0.6$ for the other networks.
On Market-1501, the hyper-parameters were set as $\alpha=1$, $\beta=0.9$ and $\gamma=1$ for all the networks.~Besides, both $\beta_1$ and $\beta_2$ were set to 2.~In all the experiments, we considered a pair of peer networks that have the same architecture or different architectures.~Table~\ref{tbbase} shows the parameter size of each network on CIFAR-100 and its top-1 accuracy on the two CIFAR datasets and the Tiny-ImageNet dataset, which serves as a baseline.

% For easily highlight the significant performance of the competing methods, the sizes of parameters needed in the peer networks and the baseline of each network on the corresponding used data sets in all the experiments are first listed in Table~\ref{tbbase}.

\subsection{Results on CIFAR-10}
\label{4.1}

Table~\ref{tb1} reports the average Top-1 accuracy of CTSL-MKT and the state-of-the-art competitors on CIFAR-10.
It is not surprising that those knowledge distillation methods
perform better than the corresponding single network due to the knowledge transfer,
except for DML and RKD with ResNet14 and ResNet18.~The possible reason for the slightly poor performance of DML and RKD with ResNet14 or ResNet18 could be that the discrepancies between outputs of small peer networks for individual instances hinder mutual learning.~Among those knowledge distillation methods, CTSL-MKT performs the best with a significant improvement,
which indicates that our idea of collaborative learning with multiple knowledge transfer is effective.~Meanwhile, CTSL-MKT outperforms all the corresponding baselines shown in Table~\ref{tbbase} with a noticeable margin.
For example, CTSL-MKT with ShuffleNetV2-MobileNetV2 has increased the Top-1 accuracy by 1.68\% and 1.18\%,
compared with the corresponding signal network baselines, {\it i.e.}, ShuffleNetV2 and MobileNetV2 respectively.~Moreover, although CTSL-MKT, DML, and RKD collaboratively learn the two peer networks, which can have
the same network structures or different ones, each peer network in CTSL-MKT ({\it i.e.}, CTSL-MKT\_$N_1$ or CTSL-MKT\_$N_2$) performs much better
than its counterpart in DML and RKD, due to the multiple knowledge transfer.

\begin{table*}[!t]%\footnotesize
\centering
\begin{adjustbox}{max width=\textwidth}
\begin{tabular}{ccccccccc}
\hline
Network $N_1$ &Network $N_2$&Tf-KD &DML\_$N_1$ &DML\_$N_2$ &RKD\_$N_1$ &RKD\_$N_2$ &CTSL-MKT\_$N_1$ &CTSL-MKT\_$N_2$  \\
\hline
ResNet14 & ResNet14&95.08$\pm$0.01 &94.78$\pm$0.02 &94.92$\pm$0.01 &94.95$\pm$0.01 &94.83$\pm$0.02  &\textbf{95.28$\pm$0.01} &95.22$\pm$0.01 \\
ResNet18 & ResNet18&95.20$\pm$0.01 &94.88$\pm$0.01 &94.99$\pm$0.01 &94.98$\pm$0.04 &94.92$\pm$0.01 &95.29$\pm$0.04 &\textbf{95.33$\pm$0.03} \\
ResNet34 &ResNet34&95.41$\pm$0.01 &95.42$\pm$0.01 &95.32$\pm$0.01 &95.45$\pm$0.01 &95.45$\pm$0.01 &\textbf{95.69$\pm$0.03} &95.59$\pm$0.01\\
MobileNetV2&MobileNetV2&91.72$\pm$0.01 &91.19$\pm$0.07 &91.32$\pm$0.04 &91.12$\pm$0.03 &90.71$\pm$0.06 &\textbf{92.12$\pm$0.02} &\textbf{92.12$\pm$0.02} \\
ShuffleNetV2&ShuffleNetV2&92.47$\pm$0.01 &91.97$\pm$0.03 &91.92$\pm$0.01 &92.08$\pm$0.01 &91.59$\pm$0.01 &\textbf{92.64$\pm$0.01} &92.49$\pm$0.01 \\
\hline
%VGG19&VGG19&87.69$\pm$0.01 &93.45$\pm$0.03 &93.36$\pm$0.01 &93.58$\pm$0.01 &93.41$\pm$0.01 &\textbf{94.04$\pm$0.01} &94.01$\pm$0.02\\
ResNet18 &ResNet34&- &95.09$\pm$0.01 &95.41$\pm$0.03 &95.12$\pm$0.01 &95.31$\pm$0.01 &95.24$\pm$0.01 &\textbf{95.60$\pm$0.01} \\
ResNet18 &VGG19&- &95.11$\pm$0.03 &93.49$\pm$0.02 &95.03$\pm$0.01 &93.50$\pm$0.01 &\textbf{95.16$\pm$0.01} &93.91$\pm$0.01 \\
ShuffleNetV2&MobileNetV2&-&91.78$\pm$0.03 &91.25$\pm$0.08 &91.73$\pm$0.01 &90.72$\pm$0.01 &\textbf{92.71$\pm$0.01} &92.15$\pm$0.01\\
\hline
\end{tabular}
\end{adjustbox}
%}
\caption{The average Top-1 accuracy (\%) over three individual runs on CIFAR-10.  }
\label{tb1}
\end{table*}

\subsection{Results on CIFAR-100}
\label{4.2}
Table~\ref{tb2} reports the average Top-1 accuracy of all the competing knowledge distillation methods with various network architectures on CIFAR-100. We have similar observations as those on CIRFAR-10.~Overall, each competing method improves on the performance of the corresponding baseline,
and CTSL-MKT gains the largest improvement.
Compare to Tf-KD, DML and RKD, Top-1 accuracy of each peer network in CTSL-MKT has been improved by about 1\% on average.~For example, the accuracy of the two MobileNetV2 networks in CTSL-MKT has been increased by 2.46\% and 2.68\% respectively,
compared to those in DML, and increased by 3.08\% and 2.96\% Top-1 accuracy compared to those in RKD.

\begin{table*}[!t]
\centering
\begin{adjustbox}{max width=\textwidth}
\begin{tabular}{ccccccccc}
\hline
Network $N_1$ &Network $N_2$&Tf-KD &DML\_$N_1$ &DML\_$N_2$ &RKD\_$N_1$ &RKD\_$N_2$ &CTSL-MKT\_$N_1$ &CTSL-MKT\_$N_2$  \\
\hline
ResNet14 & ResNet14&76.67$\pm$0.02 &75.97$\pm$0.01 &76.16$\pm$0.11 &76.36$\pm$0.03 &76.30$\pm$0.01  &\textbf{77.00$\pm$0.05} &76.85$\pm$0.04\\
ResNet18 & ResNet18&77.04$\pm$0.12 &76.10$\pm$0.10 &76.27$\pm$0.07 &76.43$\pm$0.01 &76.09$\pm$0.01 &77.43$\pm$0.10 &\textbf{77.46$\pm$0.01} \\
ResNet34 &ResNet34&77.93$\pm$0.01 &77.88$\pm$0.12 &77.61$\pm$0.03 &77.63$\pm$0.03 &77.65$\pm$0.05 &\textbf{78.58$\pm$0.01} &78.24$\pm$0.02\\
MobileNetV2&MobileNetV2&70.82$\pm$0.02 &68.98$\pm$0.01 &68.58$\pm$0.18 &68.36$\pm$0.01 &68.30$\pm$0.01 &\textbf{71.44$\pm$0.06} &71.26$\pm$0.08 \\
ShuffleNetV2&ShuffleNetV2&71.79$\pm$0.02 &70.47$\pm$0.15 &70.29$\pm$0.04 &70.24$\pm$0.01 &69.98$\pm$0.03 &\textbf{72.13$\pm$0.02} &71.69$\pm$0.05\\
\hline
ResNet18 &ResNet34&- &76.15$\pm$0.10 &77.71$\pm$0.01 &76.41$\pm$0.05 &77.83$\pm$0.01 &77.61$\pm$0.08 &\textbf{78.15$\pm$0.12}\\
ResNet18 &VGG19&- &76.51$\pm$0.02 &68.80$\pm$3.74 &76.29$\pm$0.02 &68.28$\pm$0.87 &\textbf{77.23$\pm$0.02} &72.72$\pm$0.06 \\
ShuffleNetV2&MobileNetV2&- &70.47$\pm$0.13 &68.83$\pm$0.14 &70.50$\pm$0.28 &67.87$\pm$0.01 &\textbf{72.46$\pm$0.15} &71.34$\pm$0.09\\
\hline
\end{tabular}
\end{adjustbox}
%}
\caption{The average Top-1 accuracy (\%) over three individual runs on CIFAR-100. }
\label{tb2}
\end{table*}

%\begin{figure}[!t]
%\centering
%\includegraphics[scale=0.85]{fig3_1.eps}
%\caption{The values of training loss with variations of epoch. }
%\label{fig3}
%\end{figure}

\begin{figure}[!ht]
\centering
  \subfigure[ShuffleNetV2]
  {\includegraphics[scale=1.1]{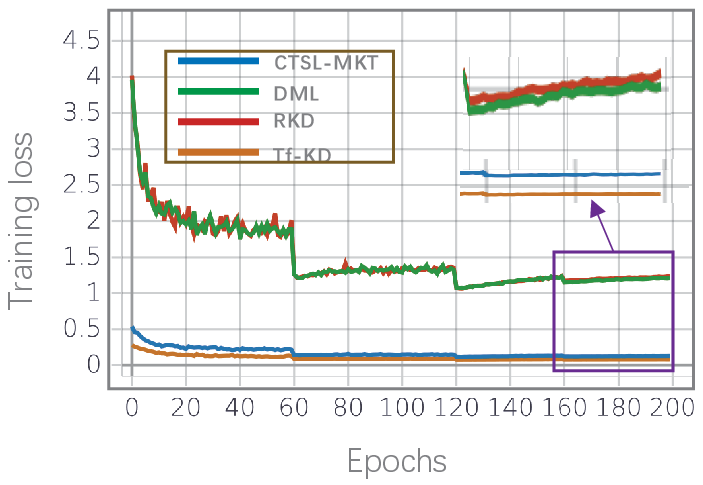}}
  \label{fig3.1}
  \subfigure[MobileNetV2]
  {\includegraphics[scale=1.1]{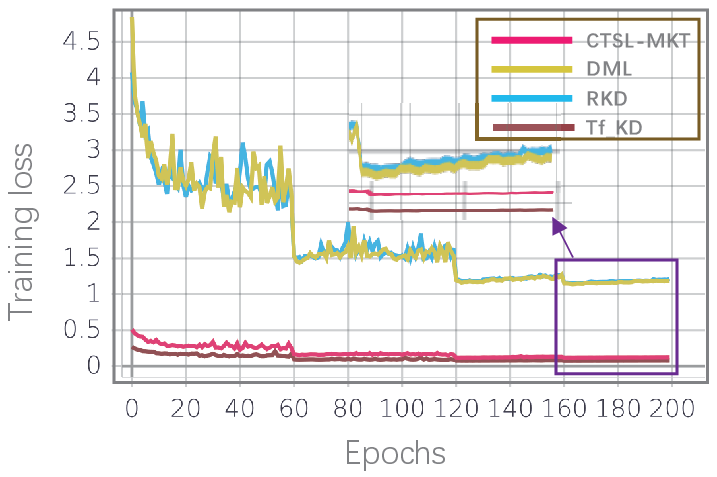}}
  \label{fig3.2}
  \caption{The values of training loss over epochs on the CIFAR-100 training dataset.}
  \label{fig3}
\end{figure}

\begin{figure}[!ht]
\centering
  \subfigure[ShuffleNetV2]
  {\includegraphics[scale=1.1]{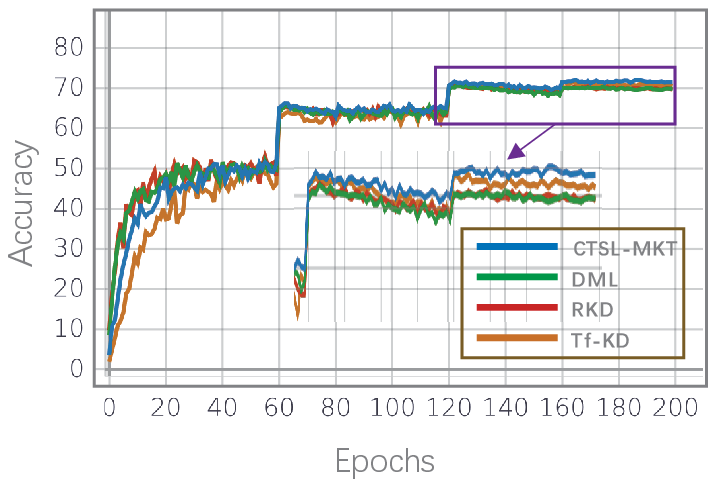}}
  \label{fig30.1}
  \subfigure[MobileNetV2]
  {\includegraphics[scale=1.1]{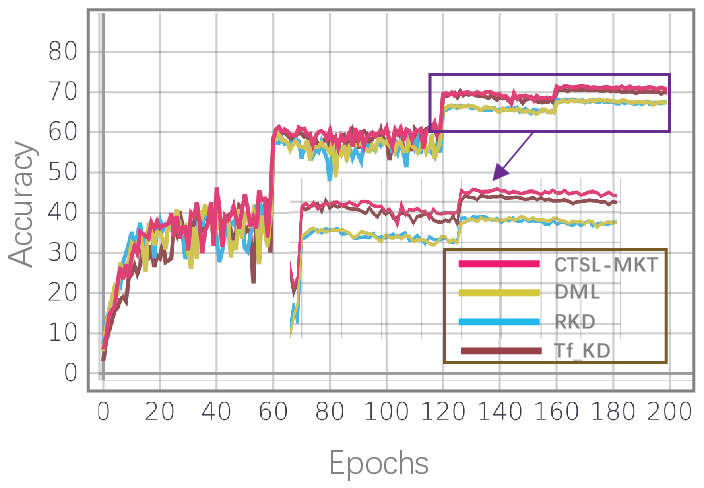}}
  \label{fig30.2}
  \caption{The values of Top-1 accuracy over epochs on the CIFAR-100 testing dataset.}
  \label{fig30}
\end{figure}

%\centering
%\includegraphics[scale=0.85]{fig3_2.eps}
%\caption{Top-1 accuracy with variations of epoch. }
%\label{fig30}
%\end{figure}

To further illustrate the learning process of the peer networks in CTSL-MKT, Figure~\ref{fig3} plots the training loss of ShuffleNetV2 and MobileNetV2 as a function of epochs, compared to Tf-KD, DML and RKD.
It shows that CTSL-MKT and Tf-KD converge better than DML and RKD.
The possible reason is that each network can self-learn in CTSL-MKT and Tf-KD to overcome the discrepancy in the outputs of peer networks in DML and RKD during learning.
Although CTSL-MKT with multiple knowledge transfer introduces extra hyper-parameters, it can still converge faster, achieving comparable training loss with Tf-KD. The loss becomes stable around 120 epochs in general.
Furthermore, Figure~\ref{fig30} displays the corresponding Top-1 accuracy of each peer network after each epoch on the testing dataset.~It shows that the proposed CTSL-MKT outperforms the others after the convergence, its performance improves along with the decrement of the training loss.~Overall, the patterns show that two peer networks in CTSL-MKT can work collaboratively, via teaching and learning from each other at each epoch, and each network gradually improves itself to achieve a better performance.

\subsection{Results on Tiny-ImageNet}
\label{4.3}
Table~\ref{tb3} shows the average Top-1 accuracy of the competing methods with five various peer network architectures on Tiny-ImageNet. From the comparative results, it can be seen that CTSL-MKT significantly outperforms the baselines, DML, RKD and Tf-KD. However, on these five peer network architectures, some peer networks in DML, RKD and Tf-KD achieve poor performance, compared to their baselines.~The possible reason is that the used peer networks are smaller with less informative knowledge in knowledge transfer and the outputs of peer networks for the same individual instances migth be different, which degrades the effectiveness of mutual learning.~With multiple kinds of knowledge and distillation strategies, our CTSL-MKT can well improve the performance via mutual learning.

\begin{table*}[!t]
\centering
\begin{adjustbox}{max width=\textwidth}
\begin{tabular}{ccccccccc}
\hline
Network $N_1$ &Network $N_2$&Tf-KD &DML\_$N_1$ &DML\_$N_2$ &RKD\_$N_1$ &RKD\_$N_2$ &CTSL-MKT\_$N_1$ &CTSL-MKT\_$N_2$  \\
\hline
ResNet18 & ResNet18&63.29$\pm$0.02 &62.30$\pm$0.01 &62.39$\pm$0.03 &62.80$\pm$0.01 &62.42$\pm$0.08 &63.63$\pm$0.08 &\textbf{63.64$\pm$0.02} \\
AlexNet &AlexNet&49.78$\pm$0.01&44.47$\pm$0.01 &44.80$\pm$0.01 &43.54$\pm$0.01 &42.97$\pm$0.01 &\textbf{51.39$\pm$0.01} &51.28$\pm$0.01\\
SqueezeNet&SqueezeNet&41.66$\pm$0.01 &47.16$\pm$0.03 &46.95$\pm$0.19 &48.22$\pm$0.01 &48.55$\pm$0.09 &48.60$\pm$0.30 &\textbf{48.86$\pm$0.03}\\
\hline
AlexNet &SqueezeNet&- &44.35$\pm$0.51 &46.15$\pm$0.30 &44.66$\pm$1.87 &46.86$\pm$0.41 &\textbf{50.98$\pm$0.08} &\textbf{47.99$\pm$0.03}\\
ResNet18 &AlexNet&- &62.62$\pm$0.11 &43.53$\pm$0.62 &62.37$\pm$0.01 &46.64$\pm$0.03 &\textbf{63.37$\pm$0.01} &\textbf{51.56$\pm$0.02}\\
\hline
\end{tabular}
\end{adjustbox}
\caption{The average Top-1 accuracy (\%) over three individual runs on Tiny-ImageNet. }
\label{tb3}
\end{table*}

\subsection{Results on Market-1501}
\label{4.4}

\begin{figure}[!t]
\centering
\includegraphics[scale=0.6]{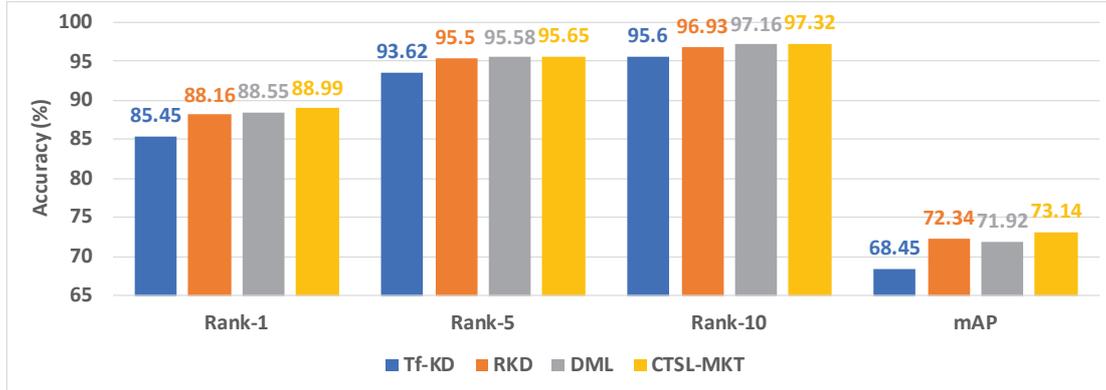}
\caption{Comparative results (\%) on Market-1501.}
\label{fig4}
\end{figure}

We further compared those methods on Market-1501, which is used for a re-identification (re-id) task.
In this set of experiments, we adopted ResNet50 that is usually used on this dataset to form a peer network architecture.~Figure~\ref{fig4} shows the performance of Tf-KD, DML, RKD and CTSL-MKT, measured by Rank-1, Rank-5, Rank-10 and mAP.
Note that these results were computed for only one peer network.
% 这个准确的说，是怎么计算的呀？特别是用到两个网络的模型
DML and RKD with collaborative learning consistently perform better than Tf-KD via self-learning.
Our CTSL-MKT outperforms both DML and RKD across all the metrics.
Specifically, in terms of mAP, the improvement of CTSL-MKT over DML, RKD and Tf-KD is 1.22\%, 0.8\% and 4.69\%, respectively.

\subsection{Ablation Study}
\label{4.5}

CTSL-MKT contains three knowledge distillation strategies,~{\it i.e.},~\textbf{m}utual \textbf{l}earning via response-based knowledge transfer from individual \textbf{i}nstances (\textbf{MLI}), \textbf{m}utual \textbf{l}earning via relation-based knowledge transfer from instance \textbf{r}elationships (\textbf{MLR}) and \textbf{s}elf-\textbf{l}earning via response-based knowledge transfer from individual \textbf{i}nstances (\textbf{SLI}).
To study how each strategy contributes to the model performance,
we consider the following four variants of CTSL-MKT:
\begin{enumerate}[itemsep=0pt,parsep=0pt,label=\emph{\Alph*})]
\item the full model using the three strategies altogether, where
we used both online distillation and self-distillation with the two kinds
of knowledge;
\item the model using online distillation only with both the response-based knowledge (MLI) and the relation-based knowledge (MLR);
\item the model using online distillation with the relation-based knowledge (MLR) and self-distillation with the response-based knowledge (SLI);
\item the model using both online distillation and self-distillation with only the response-based knowledge, corresponding to MLI + SLI.
\end{enumerate}

% Besides the experiments above of verifying the effectiveness of the proposed CTSL-MKT, we further experimentally explore the efficacy of knowledge distillation with multiple kinds of knowledge and distillation strategies ({\it i.e.} multiple knowledge transfer).
% For ablation study, four cases in CTSL-MKT are considered as follows: Furthermore, the experimental ablation study of the four cases are carried out on the various peer network architectures with the same and different structures.

Table~\ref{tb4} reports the average Top-1 accuracy of these four variations with different pairs of peer network architectures.~We have the following observations:~1) Variant A ({\it i.e.}, the full model) outperforms the other variants where one knowledge distillation strategy has been removed.~It implies that the use of multiple types of knowledge with both online distillation and self-distillation plays a crucial role in the performance gain.
2) Variant B without self-distillation has the largest performance drop, compared with variants C and D.
It indicates that self-distillation contributes substantially to the overall performance as it could offset the diversity issue caused by mutual learning and further enhance the knowledge distillation efficiency.~3) DML, Tf-KD and RKD can be seen as a special case of CTSL-MKT using only one knowledge distillation strategy.
Jointly looking at Tables~\ref{tb4} and \ref{tb2} reveals that knowledge distillation methods with two or more strategies almost always outperform those using only one strategy.~Therefore, it is clear that knowledge distillation via proper multiple knowledge transfer is very beneficial for improving the performance of model compression.

\begin{table*}[!t]
\centering
%\resizebox{.95\columnwidth}{!}{
\subtable[Ablation experiments on the same peer network architectures]{
\begin{adjustbox}{max width=\textwidth}
\begin{tabular}{c|ccc|cc|cc|cc}
\hline
    Case & MLI & MLR & SLI&ResNet14&ResNet14 &ResNet18&ResNet18&MobileNetV2&MobileNetV2\\
\hline
    A &\checkmark&\checkmark &\checkmark &\textbf{77.00$\pm$0.05}&\textbf{76.85$\pm$0.04} &\textbf{77.43$\pm$0.10} &\textbf{77.46$\pm$0.01}&\textbf{71.44$\pm$0.06}&\textbf{71.26$\pm$0.08} \\
    B & \checkmark & \checkmark & \XSolidBrush  &76.57$\pm$0.04&76.37$\pm$0.02  &76.67$\pm$0.04&76.66$\pm$0.09 &69.10$\pm$0.01&69.23$\pm$0.04\\
    C &\XSolidBrush  & \checkmark &\checkmark &76.69$\pm$0.02&76.70$\pm$0.04 &77.35$\pm$0.04 &77.29$\pm$0.08 &71.18$\pm$0.05&71.10$\pm$0.08\\
    D & \checkmark &\XSolidBrush  & \checkmark &76.73$\pm$0.03 &76.70$\pm$0.08 &77.26$\pm$0.07&77.12$\pm$0.04 &71.14$\pm$0.02&71.04$\pm$0.10\\
\hline
\end{tabular}
\end{adjustbox}
\label{tb4.1}
}
\quad
\subtable[Ablation experiments on the different peer network architectures]{
\begin{adjustbox}{max width=\textwidth}
\begin{tabular}{c|ccc|cc|cc|cc}
\hline
    Case & MLI & MLR & SLI&ResNet14&ResNet18 &ResNet18&ResNet34 &ShuffleNetV2&MobileNetV2\\
\hline
    A &\checkmark&\checkmark &\checkmark &\textbf{77.07$\pm$0.03}&\textbf{77.28$\pm$0.04} & \textbf{77.61$\pm$0.08}&\textbf{78.15$\pm$0.12}&\textbf{72.46$\pm$0.15}&\textbf{71.34$\pm$0.09} \\
    B & \checkmark & \checkmark  &\XSolidBrush  &76.35$\pm$0.01 &76.53$\pm$0.07&76.53$\pm$0.02  &77.83$\pm$0.01 &70.57$\pm$0.03&68.69$\pm$0.01\\
    C &\XSolidBrush  & \checkmark &\checkmark &76.69$\pm$0.23&77.06$\pm$0.02 &77.12$\pm$0.01 &77.99$\pm$0.03 &72.06$\pm$0.04&71.10$\pm$0.11\\
    D & \checkmark &\XSolidBrush  & \checkmark &76.68$\pm$0.03&77.13$\pm$0.02 &77.39$\pm$0.01&77.68$\pm$0.01 &72.20$\pm$0.09&71.05$\pm$0.15\\
\hline
\end{tabular}
\end{adjustbox}
\label{tb4.2}
}
\caption{Ablation study of CTSL-MKT in terms of the average Top-1 accuracy over three individual runs on CIFAR-100. }
\label{tb4}
\end{table*}

\subsection{Experiment Discussion}
\label{4.6}
The experimental results reported above have demonstrated the effectiveness of the proposed CTSL-MKT, while being
compared with several state-of-the-art knowledge distillation methods. We have the following remarks:
\begin{itemize}[itemsep=0pt,parsep=0pt]
\item Collaborative learning can make peer networks teach and learn from each other, and iteratively improve themselves.
\item Self-learning of each peer network can further enhance the ability of mutual learning among peer networks by compensating the loss caused by the diversity issue.
\item Multiple knowledge transfer with more than one types of knowledge and distillation strategies can significantly improve the KD performance.
\item Various peer network architectures ({\it i.e.},~teacher-student architectures) can be easily adopted for knowledge transfer via collaborative learning.
\end{itemize}

\section{Conclusions}

In this paper, we propose a novel knowledge distillation method called collaborative teacher-student learning via multiple knowledge transfer (CTSL-MKT). It naturally integrates both self-learning via self-distillation and collaborative learning via online distillation in a unified framework so that multiple kinds of knowledge can be transferred effectively in-between different teacher-student architectures,
and CTSL-MKT can achieve individual instance consistency and instance correlation consistency among the peer networks.~Experimental results on four image classification datasets
have demonstrated CTSL-MKT outperforms the competitors with a noticeable margin, which proves the necessity of using different distillation schemes to transfer multiple types of knowledge simultaneously.~We believe that our proposed framework opens a door to design multiple knowledge transfer for knowledge distillation.

%{\color{red}In future work, we would like to XXXXX}

%\section*{Acknowledgment}
%
%This work was supported by Postgraduate Research \& Practice Innovation Program of Jiangsu Province (No. KYCX20\_3085).



%% References


\bibliographystyle{model1a-num-names}
\bibliography{<your-bib-database>}



\end{document}